\documentclass{article}
\usepackage[preprint]{neurips_2025}

\usepackage[utf8]{inputenc}
\usepackage[T1]{fontenc}
\usepackage{hyperref}
\usepackage{url}
\usepackage{booktabs}
\usepackage{amsfonts}
\usepackage{amsmath}
\usepackage{nicefrac}
\usepackage{microtype}
\usepackage{xcolor}
\usepackage{graphicx}
\usepackage{subcaption}
\usepackage{longtable}
\usepackage{array}
\usepackage{multirow}
\usepackage{adjustbox}
\usepackage{float}

\title{KV Cache Quantization for Self-Forcing Video Generation:\\A 33-Method Empirical Study}

\author{%
  Suraj Ranganath\thanks{\texttt{suranganath@ucsd.edu}.} \\
  University of California, San Diego
  \And
  Vaishak Menon \\
  University of California, San Diego
  \And
  Anish Patnaik \\
  University of California, San Diego
}

\newcommand{\method}[1]{\texttt{#1}}
\newcommand{\methodstack}[2]{\shortstack[l]{\method{#1}\\\method{#2}}}
\newcommand{\methodstackthree}[3]{\shortstack[l]{\method{#1}\\\method{#2}\\\method{#3}}}
\newcommand{\benchmark}[1]{\textsc{#1}}

\begin{document}

\maketitle

\begin{abstract}
Self-forcing video generation extends a short-horizon video model to longer rollouts by repeatedly feeding generated content back in as context. This scaling path immediately exposes a systems bottleneck: the key-value (KV) cache grows with rollout length, so longer videos require not only better generation quality but also substantially better memory behavior. We present a comprehensive empirical study of KV-cache compression for self-forcing video generation on a \method{Wan2.1}-based Self-Forcing stack. Our study covers 33 quantization and cache-policy variants, 610 prompt-level observations, and 63 benchmark-level summaries across two evaluation settings: \benchmark{MovieGen} for single-shot 10-second generation and \benchmark{StoryEval} for longer narrative-style stability. We jointly evaluate peak VRAM, runtime, realized compression ratio, VBench imaging quality, BF16-referenced fidelity (SSIM, LPIPS, PSNR), and terminal drift. Three findings are robust. First, the strongest practical operating region is a FlowCache-inspired soft-prune INT4 adaptation, which reaches \mbox{$5.42$--$5.49\times$} compression while reducing peak VRAM from 19.28\,GB to about 11.7\,GB with only modest runtime overhead. Second, the highest-fidelity compressed methods, especially \method{PRQ\_INT4} and \method{QUAROT\_KV\_INT4}, are not the best deployment choices because they preserve quality at severe runtime or memory cost. Third, nominal compression alone is not sufficient: several methods shrink KV storage but still exceed BF16 peak VRAM because the current integration reconstructs or retains large BF16 buffers during attention and refresh stages. The result is a benchmark harness, analysis workflow, and empirical map of which KV-cache ideas are practical today and which are promising research directions for better memory integration. Code, data products, and the presentation dashboard are available at \url{https://github.com/suraj-ranganath/kv-quant-longhorizon/}.
\end{abstract}

\section{Introduction}

Autoregressive video diffusion has become a compelling route to long-horizon video generation because it can roll forward indefinitely by treating previously generated content as context for future frames. Self-Forcing makes this idea especially practical by training with self-generated context and KV-cache rollouts rather than relying only on teacher-forced conditioning \citep{huang2025selfforcing}. The same mechanism that makes long rollout possible also creates a central scaling bottleneck: as context grows, the KV cache grows, and that cache must be stored, read, written, and sometimes transformed repeatedly during generation. In other words, longer rollout is not limited only by model quality. It is also limited by memory capacity, memory bandwidth, and the implementation details of how cached state is represented.

This observation motivates a specific empirical question. If we compress the KV cache, which methods actually improve end-to-end long-video generation, and which ones merely look efficient in isolation? A method can report an attractive nominal compression ratio while still failing in deployment because it introduces large dequantization buffers, refresh overhead, or quality collapse under rollout. For video generation, these systems and quality axes cannot be separated cleanly. A useful method must reduce memory pressure enough to matter, remain computationally viable, preserve perceptual realism, and avoid structural drift relative to the uncompressed reference.

We study this question on a \method{Wan2.1}-based Self-Forcing stack \citep{wan2025wan,huang2025selfforcing}. We benchmark 33 methods spanning naive quantization, asymmetric quantization, rotation-based quantization, progressive residual coding, temporal heuristics, spatial mixed precision, and FlowCache-inspired retention/pruning policies. We evaluate both single-shot video quality and narrative-style rollout stability, and we aggregate each method through six axes: peak VRAM, runtime, KV compression ratio, VBench imaging quality \citep{huang2023vbench}, BF16-relative fidelity, and drift-last imaging quality. Our dataset comprises 610 prompt-level observations and 63 benchmark-level method summaries.

The main empirical result is not that one universal method dominates every objective. Instead, the study reveals a structured design space. FlowCache-inspired soft pruning is the strongest practical operating point for real memory relief. \method{PRQ\_INT4} and \method{QUAROT\_KV\_INT4} define a high-fidelity region that is academically useful but operationally expensive. Recency- and refresh-based RTN variants reveal policy ideas that help quality, yet their current integration still allows transient BF16 reconstruction to dominate peak VRAM. Spatially mixed methods demonstrate a clear negative result: a plausible spatial-importance heuristic can fail catastrophically for autoregressive video generation.

\section{Related Work}

Our work sits at the intersection of modern video generation, autoregressive long-horizon rollout, and KV-cache systems optimization. Large-scale video foundation models such as Movie Gen and Wan demonstrate that transformer-based video generation can achieve strong perceptual quality when given enough scale, context length, and inference engineering \citep{polyak2024moviegen,wan2025wan}. Self-Forcing extends this trajectory by explicitly addressing the train-test mismatch in autoregressive video diffusion and by making KV-cached rollout part of training and inference rather than a purely downstream systems optimization \citep{huang2025selfforcing}.

In parallel, the systems community has shown that KV-cache management is a first-order bottleneck in large-sequence inference. PagedAttention made this point explicit for language models by showing that request-level throughput can be constrained as much by KV memory management as by raw arithmetic throughput \citep{kwon2023pagedattention}. This insight transfers naturally to long-horizon video generation, where cache growth is tied directly to time. FlowCache extends the idea from generic memory management to autoregressive video generation by proposing chunkwise caching policies and KV compression mechanisms tailored to chunked video rollout \citep{ma2026flowcache}. Our study is closely related in spirit, but the official FlowCache implementation does not support the Self-Forcing \method{Wan2.1} stack used here, so the FlowCache-labeled rows in our experiments should be read as an in-house adaptation of FlowCache-inspired ideas rather than an upstream reproduction.

The quantization literature gives several concrete tools for compressing KV tensors. KIVI uses asymmetric statistics-aware quantization for keys and values and is a strong low-bit reference for cache compression \citep{liu2024kivi}. QuaRot attacks the outlier problem directly by rotating activations into a more quantization-friendly basis before low-bit coding \citep{ashkboos2024quarot}. More recently, Quant VideoGen (QVG) proposes a training-free KV-cache quantization framework for autoregressive video generation and reports strong memory-quality Pareto improvements, including a higher-quality operating mode denoted QVG-Pro \citep{xi2026qvg}. Our benchmark includes these methods together with project-specific extensions: PRQ, QAQ, Age-Tier, TPTQ, FlowCache-inspired variants, and spatial mixed-precision policies. The goal is not to introduce a single new quantizer in isolation, but to compare a wide family of plausible design ideas under one end-to-end video-generation harness.

Recent long-video systems also provide important complementary context. CausVid distills a slow bidirectional teacher into a fast causal student for on-the-fly generation \citep{yin2025causvid}, while HiAR proposes hierarchical denoising to reduce error accumulation in long autoregressive rollouts \citep{zou2026hiar}. These systems target long-horizon generation from different angles than direct KV compression, which is exactly why they matter for future benchmarking: if our conclusions transfer across these causal generators, then the lessons are likely about cache policy itself rather than only about one implementation of Self-Forcing.

Finally, evaluation itself is a nontrivial challenge for video generation. VBench provides a multi-dimensional realism benchmark and has become a practical tool for separating perceptual quality from other axes of model behavior \citep{huang2023vbench}. In this project, we complement VBench with BF16-referenced SSIM, LPIPS, and PSNR, and with prefix-based drift curves. This dual-axis view is important because some methods remain visually plausible while structurally diverging from the BF16 baseline.

\section{Method Families and Benchmark Harness}

\subsection{Problem Setting}

We study KV-cache compression for a Self-Forcing video generator built on top of \method{Wan2.1}-style causal video generation. Each rollout generates 165 frames at 480\,$\times$\,832 and 16\,fps, corresponding to approximately 10.3 seconds of video. At this horizon, the cache is already large enough that memory-management choices materially affect feasibility in single-GPU settings. We therefore treat KV compression as a multi-objective optimization problem over memory, runtime, perceptual realism, structural fidelity, and temporal stability.

\subsection{Benchmarks}

We evaluate two complementary settings. \benchmark{MovieGen} is a single-shot prompt benchmark derived from a MovieGen-style prompt suite and is designed to reveal systems behavior and prompt-level output quality in isolated 10-second samples. \benchmark{StoryEval} uses 10 narrative-style prompts and evaluates whether a method remains stable over a longer causal rollout, summarized through a drift curve over prefix duration and a drift-last imaging-quality score. Together, these benchmarks separate immediate per-prompt quality from long-horizon stability under self-conditioning.

\subsection{Method Families}

Across the full study we evaluate 33 methods. Table~\ref{tab:families} enumerates the full inventory and groups the variants by design idea. BF16 is the reference operating point. RTN and KIVI are standard low-bit baselines. QuaRot, PRQ, and QAQ aim to preserve fidelity more carefully. Age-Tier and TPTQ introduce recency-aware temporal heuristics. The FlowCache-inspired methods move beyond plain quantization and ask which cache chunks should be retained, compressed, summarized, or reused. Spatial mixed-precision variants test whether foreground/background partitioning is a useful inductive bias for autoregressive rollout. Appendix~A provides code-level implementation notes for every evaluated method.

\begin{table*}[t]
\centering
\tiny
\setlength{\tabcolsep}{3pt}
\begin{adjustbox}{width=\textwidth}
\begin{tabular}{>{\raggedright\arraybackslash}p{1.9cm} >{\raggedright\arraybackslash}p{5.7cm} >{\raggedright\arraybackslash}p{6.8cm} >{\raggedright\arraybackslash}p{2.6cm}}
\toprule
Family & Exact methods in this study & What changes across variants & Provenance / role \\
\midrule
BF16 & \method{BF16} & Uncompressed reference cache stored in native BF16 precision; all BF16-relative fidelity deltas are measured against this baseline. & Native model baseline. \\
RTN & \method{RTN\_INT2}, \method{RTN\_INT4}, \method{RTN\_K2\_V4}, \\ \method{RTN\_INT4\_RECENT2}, \method{RTN\_INT4\_REFRESH} & INT2 and INT4 set uniform 2- or 4-bit block quantization; K2\_V4 uses 2-bit keys and 4-bit values; Recent2 preserves a recent BF16 context window; Refresh periodically re-quantizes the cache to limit drift. & Standard low-bit baseline; custom video-stack integration. \\
KIVI & \method{KIVI\_INT2}, \method{KIVI\_INT4}, \method{KIVI\_K2\_V4}, \\ \method{KIVI\_INT4\_REFRESH} & Uses asymmetric channel-wise key coding and token-wise value coding; the variants mirror RTN bit-widths, asymmetric K/V precision, and refresh cadence. & KIVI baseline \citep{liu2024kivi}. \\
QuaRot & \method{QUAROT\_KV\_INT2}, \method{QUAROT\_KV\_INT4}, \\ \methodstack{QUAROT\_KV\_INT4\_}{RECENT2}, \\ \methodstack{QUAROT\_KV\_INT4\_}{REFRESH} & Adds Hadamard rotation before RTN-style coding to suppress outliers; the recent-window and refresh variants test the same heuristics after rotation. & QuaRot baseline \citep{ashkboos2024quarot}. \\
\shortstack[l]{Project high-fidelity\\quantizers} & \method{PRQ\_INT2}, \method{PRQ\_INT4}, \method{QAQ\_INT2}, \method{QAQ\_INT4} & PRQ uses two-stage progressive residual coding; QAQ stores large outliers separately from the clipped low-bit bulk; INT2 and INT4 set the base storage budget. & Novel methods in this project. \\
\shortstack[l]{Temporal\\heuristics} & \method{AGE\_TIER\_INT2}, \method{AGE\_TIER\_INT4}, \method{TPTQ\_INT2} & Age-Tier keeps recent tokens at higher precision while compressing older history more aggressively; TPTQ adds a PRQ-coded older zone plus explicit outlier preservation. & Novel methods in this project. \\
\shortstack[l]{FlowCache-inspired\\adaptations} & \methodstack{FLOWCACHE\_}{HYBRID\_INT2}, \\ \methodstack{FLOWCACHE\_}{ADAPTIVE\_INT2}, \\ \methodstack{FLOWCACHE\_}{PRUNE\_INT2}, \methodstack{FLOWCACHE\_}{PRUNE\_INT4}, \\ \methodstack{FLOWCACHE\_SOFT\_}{PRUNE\_INT2}, \methodstack{FLOWCACHE\_SOFT\_}{PRUNE\_INT4}, \\ \method{FLOWCACHE\_NATIVE}, \\ \methodstackthree{FLOWCACHE\_}{NATIVE\_SOFT\_}{PRUNE\_INT4} & Hybrid uses chunk-age and layer-aware budgets; Adaptive adds drift-aware allocation; Prune hard-evicts low-importance chunks; Soft-Prune replaces evicted chunks with summaries; Native reuses internal features; Native-Soft-Prune combines reuse with soft-pruned INT4 caching. & In-house FlowCache-style adaptation \citep{ma2026flowcache}; not a direct upstream reproduction on \method{Wan2.1}. \\
\shortstack[l]{Spatial mixed\\precision} & \methodstackthree{SPATIAL\_MIXED\_FG\_}{RTN\_INT4\_BG\_}{RTN\_INT2}, \\ \methodstackthree{SPATIAL\_MIXED\_FG\_}{RTN\_INT4\_BG\_}{RTN\_INT4}, \\ \methodstackthree{SPATIAL\_MIXED\_FG\_}{KIVI\_INT4\_BG\_}{KIVI\_INT2}, \\ \methodstackthree{SPATIAL\_MIXED\_FG\_}{QUAROT\_KV\_INT4\_BG\_}{RTN\_INT2} & Uses a motion-derived spatial mask so foreground tokens receive a higher-fidelity encoder and background tokens a cheaper one; the four variants test RTN-only, KIVI-only, and QuaRot-foreground mixes. & Novel methods in this project. \\
\bottomrule
\end{tabular}
\end{adjustbox}
\caption{Complete method inventory for the 33-method study. The table lists every evaluated variant and clarifies what changes within each family. Appendix~A provides code-level implementation notes for each method.}
\label{tab:families}
\end{table*}

\subsection{Evaluation Axes}

Every method is evaluated along six axes. Peak VRAM is the maximum allocated GPU memory observed in generation logs. Runtime is the end-to-end wall-clock generation time per prompt. Compression ratio is the BF16-equivalent KV footprint divided by the compressed KV footprint. Perceptual realism is primarily measured through VBench imaging quality \citep{huang2023vbench}. Structural fidelity is measured against BF16 references using SSIM, LPIPS, and PSNR from exact video comparisons. Temporal stability is summarized by drift-last imaging quality, the final point on the prefix-quality curve.

This metric design is deliberate and follows a dual-axis view of quality. VBench answers whether a video still looks plausible as a video. SSIM, LPIPS, and PSNR answer whether it stayed faithful to the BF16 reference without hallucinating away scene structure. Prefix drift answers whether the method remains stable as self-generated context accumulates. In other words, the benchmark is a multi-axis gauntlet over memory, runtime, compression, realism, fidelity, and drift rather than a single-score leaderboard. We report PSNR for completeness, but because BF16 is compared against itself, BF16 PSNR is infinite by definition and some compressed methods also inherit non-finite aggregates if a subset of frames match exactly. For that reason, our decision analysis primarily emphasizes SSIM and LPIPS rather than PSNR alone.

\section{Experimental Setup}

All methods are run through a common harness. Video generation is launched from a single generation entrypoint that attaches an optional quantizer at the causal KV-cache boundary, logs runtime and peak VRAM, writes out prompt-level videos, and records per-prompt traces. Fidelity is then computed against BF16 reference videos by exact frame alignment with PSNR, SSIM, and LPIPS. VBench evaluation is run separately. For \benchmark{StoryEval}, a dedicated script truncates generated videos to a sequence of prefixes and re-runs VBench imaging quality over each prefix to obtain a drift curve.

The final combined comparison dataset contains 610 prompt-level rows and aggregates to 63 benchmark-level method summaries. Each summary contains method metadata, systems metrics, VBench realism scores, BF16-referenced fidelity, and drift statistics. All paper tables and figures are generated directly from that dataset or from the original trace logs. This is the same data model used by the accompanying Streamlit dashboard, so the manuscript, plots, and interactive presentation view all derive from a single source of truth. Appendix~B collects the additional repository-aligned figures, and Appendix~C includes full benchmark tables for both benchmarks.

\section{Results}

\subsection{Global Trade-offs Across 33 Methods}

\begin{figure*}[t]
  \centering
  \includegraphics[width=0.9\textwidth]{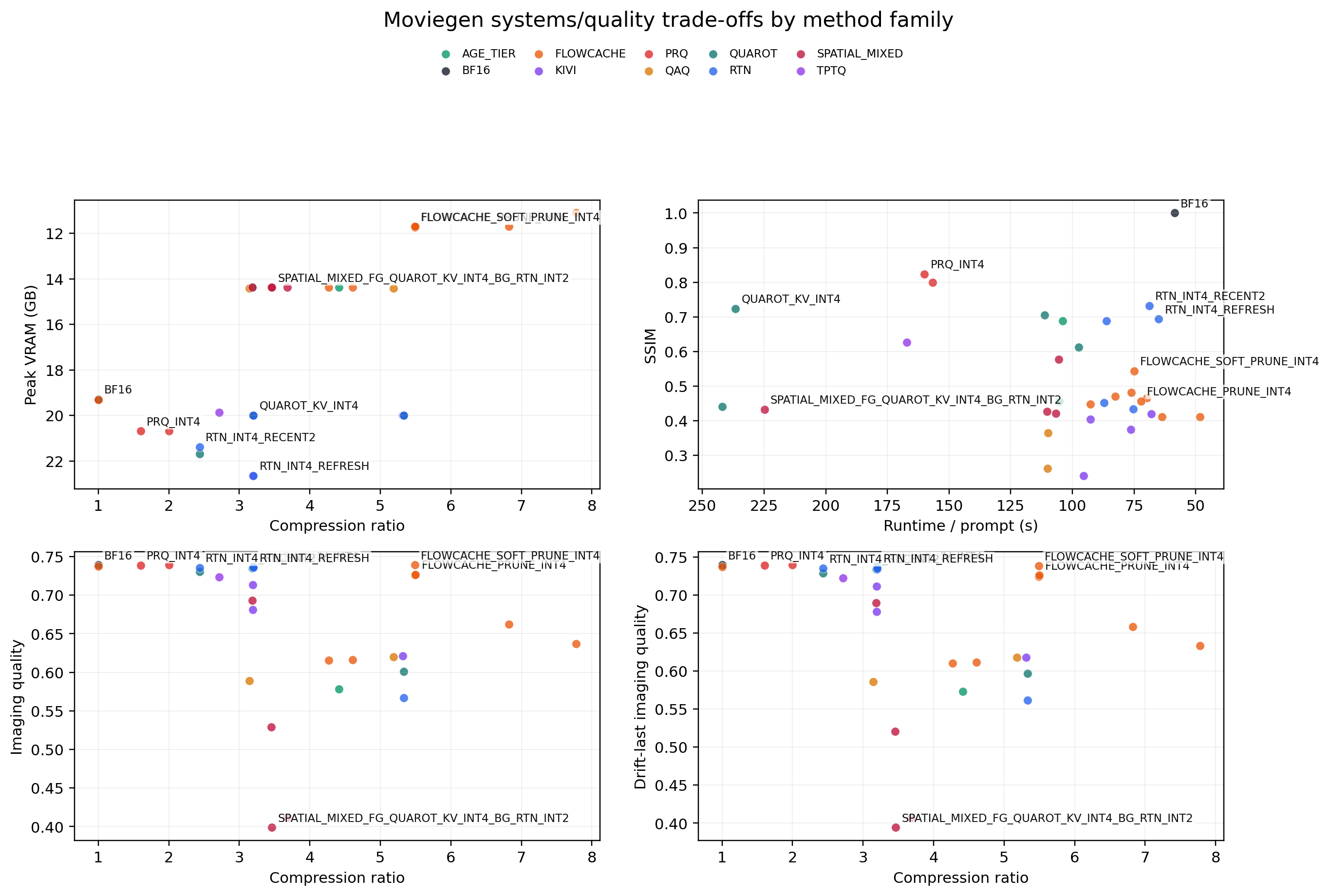}
  \caption{Global systems-quality landscape on \benchmark{MovieGen}. FlowCache-inspired prune/soft-prune methods dominate the practical low-VRAM region, PRQ and QuaRot occupy the high-fidelity region, and spatially mixed methods collapse despite plausible motivation.}
  \label{fig:tradeoffs-moviegen}
\end{figure*}

\begin{figure*}[t]
  \centering
  \includegraphics[width=0.9\textwidth]{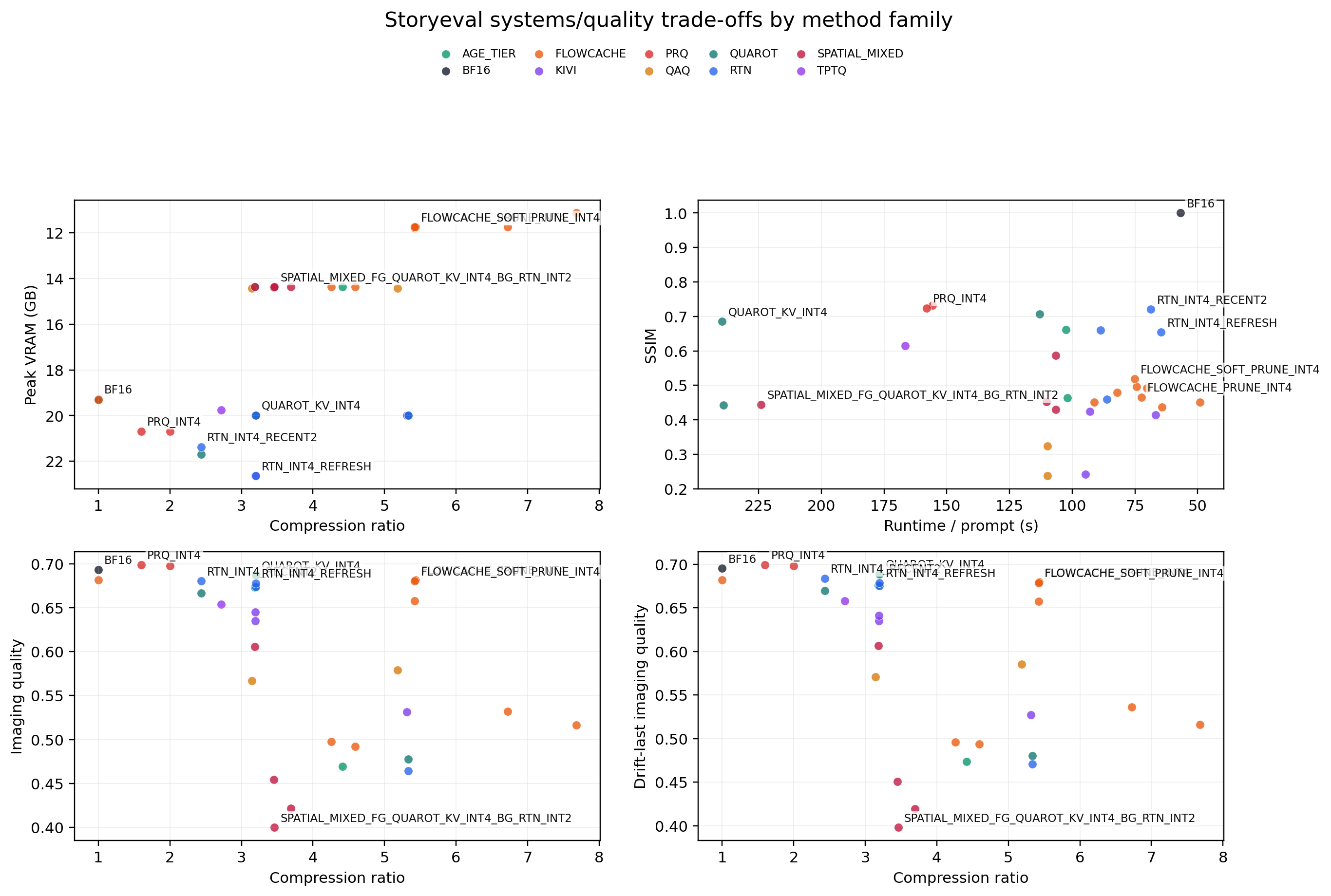}
  \caption{Global systems-quality landscape on \benchmark{StoryEval}. The qualitative structure mirrors \benchmark{MovieGen}: the practical winner remains in the FlowCache-inspired soft-prune region, while the highest-fidelity compressed methods sit at much higher runtime or peak-memory cost.}
  \label{fig:tradeoffs-storyeval}
\end{figure*}

Table~\ref{tab:main-results} summarizes the main archetypal methods. Figures~\ref{fig:tradeoffs-moviegen} and~\ref{fig:tradeoffs-storyeval} show the full trade-off landscape for all methods on both benchmarks. Several patterns appear immediately.

First, naive compression is not enough. \method{RTN\_INT4} reduces nominal KV size by $3.20\times$ on both benchmarks, yet peak VRAM remains essentially unchanged relative to BF16: 19.98\,GB versus 19.28\,GB on both \benchmark{MovieGen} and \benchmark{StoryEval}. It also degrades structural fidelity substantially, reaching SSIM 0.688 on \benchmark{MovieGen} and 0.661 on \benchmark{StoryEval}. This is the clearest evidence that nominal low-bit coding alone does not solve the deployment problem for self-forcing video generation.

Second, the strongest high-fidelity methods are not automatically the best systems choices. \method{PRQ\_INT4} is one of the closest compressed methods to BF16 on structural fidelity: on \benchmark{MovieGen} it reaches SSIM 0.824 and LPIPS 0.082, and on \benchmark{StoryEval} it reaches SSIM 0.724 with the highest drift-last quality among non-BF16 methods. But it is systems-negative, taking about 160\,s per prompt and peaking at 20.69\,GB. \method{QUAROT\_KV\_INT4} occupies a similar region: better quality than plain RTN, but 236--240\,s runtime and peak VRAM still near 20\,GB.

Third, the strongest practical memory-quality operating point is a FlowCache-inspired soft-prune INT4 adaptation. On \benchmark{MovieGen}, \method{FLOWCACHE\_SOFT\_PRUNE\_INT4} reaches $5.49\times$ compression, 11.71\,GB peak VRAM, 75.0\,s runtime, and imaging quality 0.739, effectively matching BF16 on perceptual realism while preserving drift-last quality at 0.738. On \benchmark{StoryEval}, it reaches $5.42\times$ compression, 11.76\,GB peak VRAM, and drift-last quality 0.679. This is the best practical single-GPU point in the current stack.

The most important caveat is that this practical winner is not the strongest BF16-faithfulness winner. Its VBench realism remains near BF16, but its SSIM and LPIPS are noticeably worse than PRQ or QuaRot. This ``FlowCache paradox'' is scientifically useful rather than embarrassing: it shows that perceptual plausibility and exact structural fidelity can separate sharply in long-horizon video generation.

\begin{table*}[t]
\centering
\scriptsize
\begin{tabular}{l>{\raggedright\arraybackslash}p{4.4cm}rrrrrr}
\toprule
Benchmark & Method & Comp. & Peak VRAM & Runtime & Img. & SSIM & Drift \\
\midrule
Moviegen & BF16 & 1.00 & 19.28 & 58.6 & 0.739 & 1.000 & 0.739 \\
Moviegen & FlowCache Soft-Prune INT4 & 5.49 & 11.71 & 75.0 & 0.739 & 0.544 & 0.738 \\
Moviegen & FlowCache Prune INT4 & 5.50 & 11.71 & 72.2 & 0.727 & 0.457 & 0.726 \\
Moviegen & PRQ INT4 & 1.60 & 20.69 & 160.0 & 0.739 & 0.824 & 0.739 \\
Moviegen & QuaRot KV INT4 & 3.20 & 19.98 & 236.6 & 0.738 & 0.724 & 0.738 \\
Moviegen & RTN INT4 Recent2 & 2.43 & 21.37 & 68.9 & 0.736 & 0.732 & 0.735 \\
Moviegen & RTN INT4 Refresh & 3.20 & 22.64 & 65.0 & 0.736 & 0.693 & 0.735 \\
Moviegen & Spatial Mixed (QuaRot fg, RTN bg) & 3.46 & 14.38 & 224.8 & 0.399 & 0.433 & 0.394 \\
Storyeval & BF16 & 1.00 & 19.28 & 56.8 & 0.693 & 1.000 & 0.695 \\
Storyeval & FlowCache Soft-Prune INT4 & 5.42 & 11.76 & 75.2 & 0.680 & 0.518 & 0.679 \\
Storyeval & FlowCache Prune INT4 & 5.43 & 11.75 & 72.4 & 0.682 & 0.465 & 0.680 \\
Storyeval & PRQ INT4 & 1.60 & 20.69 & 158.0 & 0.699 & 0.724 & 0.699 \\
Storyeval & QuaRot KV INT4 & 3.20 & 19.98 & 239.6 & 0.687 & 0.685 & 0.689 \\
Storyeval & RTN INT4 Recent2 & 2.43 & 21.37 & 68.6 & 0.680 & 0.721 & 0.684 \\
Storyeval & RTN INT4 Refresh & 3.20 & 22.64 & 64.6 & 0.678 & 0.654 & 0.679 \\
Storyeval & Spatial Mixed (QuaRot fg, RTN bg) & 3.46 & 14.38 & 224.1 & 0.400 & 0.444 & 0.398 \\
\bottomrule
\end{tabular}
\caption{Representative operating points across both benchmarks. \method{FLOWCACHE\_SOFT\_PRUNE\_INT4} is the strongest practical deployment choice, whereas \method{PRQ\_INT4} and \method{QUAROT\_KV\_INT4} define a high-fidelity but systems-expensive region.}
\label{tab:main-results}
\end{table*}

\subsection{Pareto and Frontier Analysis}

\begin{figure*}[t]
  \centering
  \includegraphics[width=0.9\textwidth]{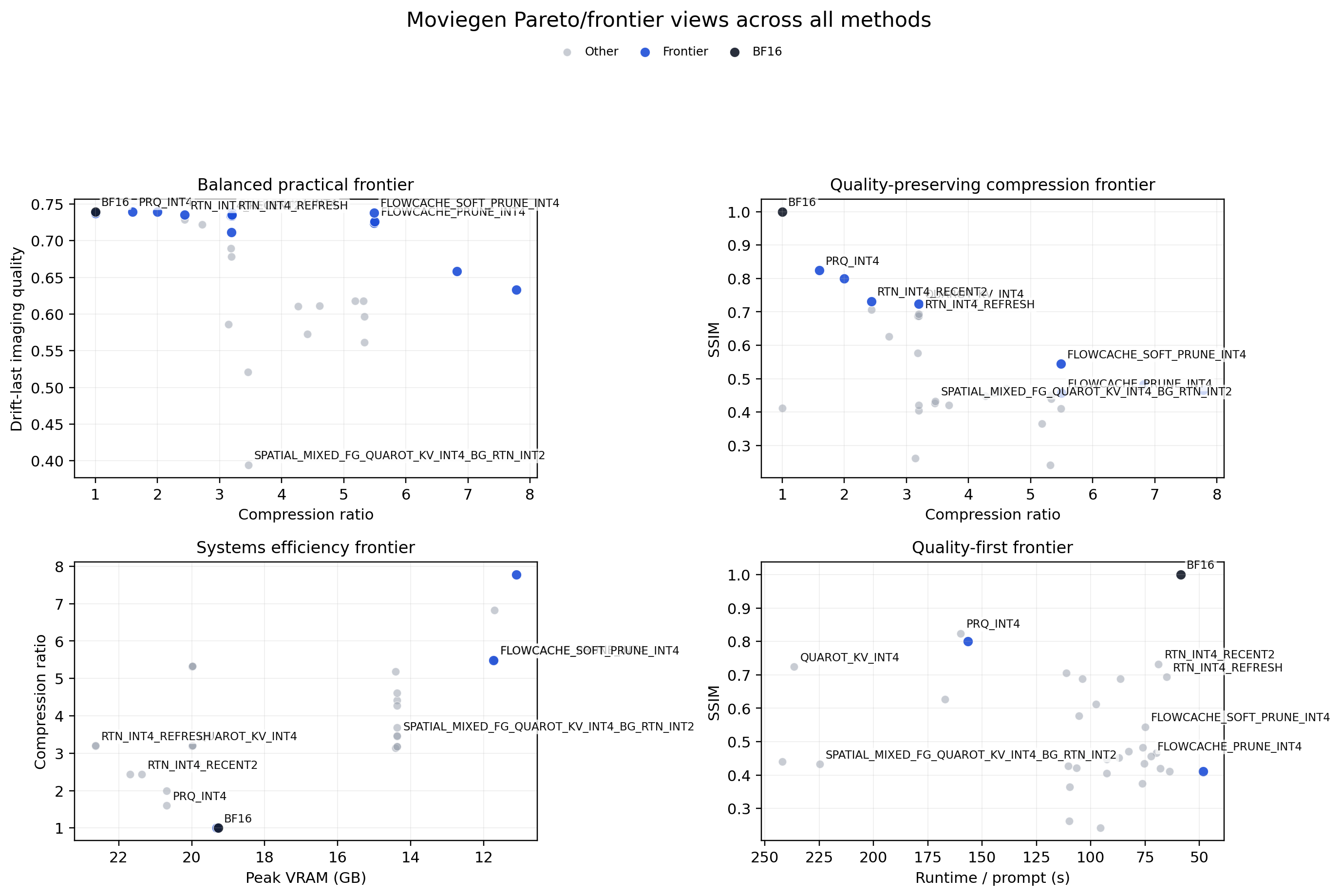}
  \caption{Pareto/frontier analysis on \benchmark{MovieGen}. Methods that survive the quality-preserving compression frontier are not the same methods that survive the systems-efficiency frontier, which is why deployment winners and research-quality winners diverge.}
  \label{fig:frontiers-moviegen}
\end{figure*}

\begin{figure*}[t]
  \centering
  \includegraphics[width=0.9\textwidth]{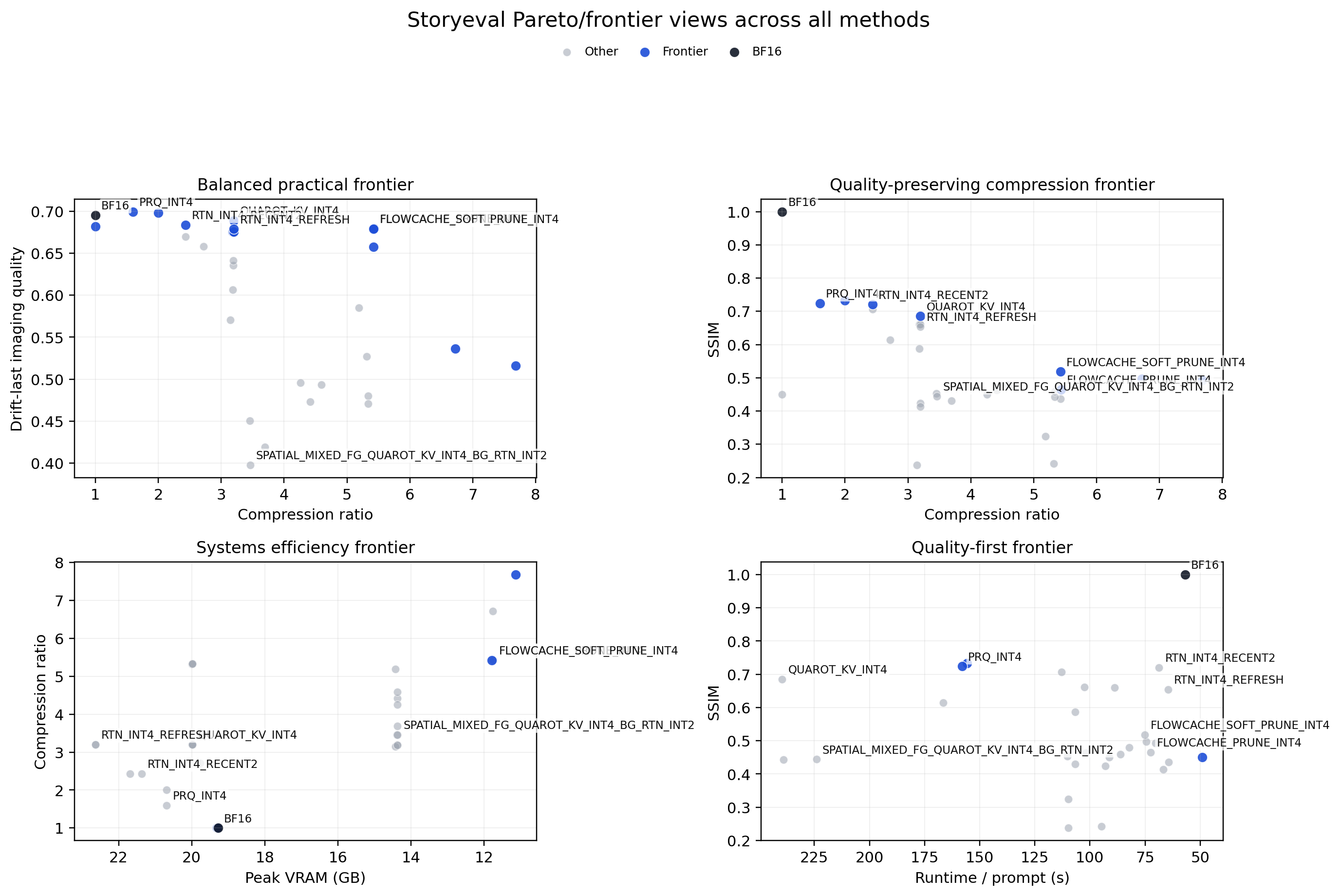}
  \caption{Pareto/frontier analysis on \benchmark{StoryEval}. The same separation persists under narrative-style rollout stability, reinforcing that the design-space conclusions are not artifacts of one benchmark.}
  \label{fig:frontiers-storyeval}
\end{figure*}

A single scalar ranking obscures too much structure, so we also analyze four frontiers: balanced practical, quality-preserving compression, systems efficiency, and quality-first. Figures~\ref{fig:frontiers-moviegen} and~\ref{fig:frontiers-storyeval} show these frontiers for all methods on both benchmarks. The frontier picture is stable across benchmarks. Compression-oriented frontiers consistently include the INT4 FlowCache-inspired prune and soft-prune variants, while quality-oriented frontiers include BF16, PRQ, QuaRot, and recency-aware RTN variants. Systems-efficiency frontiers are much smaller, reflecting the fact that only a few methods truly improve memory without paying catastrophic runtime or quality cost.

The cross-benchmark agreement is unusually strong. When we correlate benchmark-level summaries across common methods, compression ratio correlates at $r=0.9999$, runtime at $r=0.9996$, peak VRAM at $r=0.99999$, imaging quality at $r=0.9318$, and drift-last quality at $r=0.9374$. This matters because it suggests that the high-level conclusions are not an artifact of one prompt suite. In particular, the practical recommendation does not flip between \benchmark{MovieGen} and \benchmark{StoryEval}: FlowCache-inspired soft pruning remains the best operational compromise, while PRQ and QuaRot remain quality-first reference points.

\clearpage
\subsection{Qualitative Behavior}

\begin{figure*}[t]
  \centering
  \includegraphics[width=\textwidth,height=0.62\textheight,keepaspectratio]{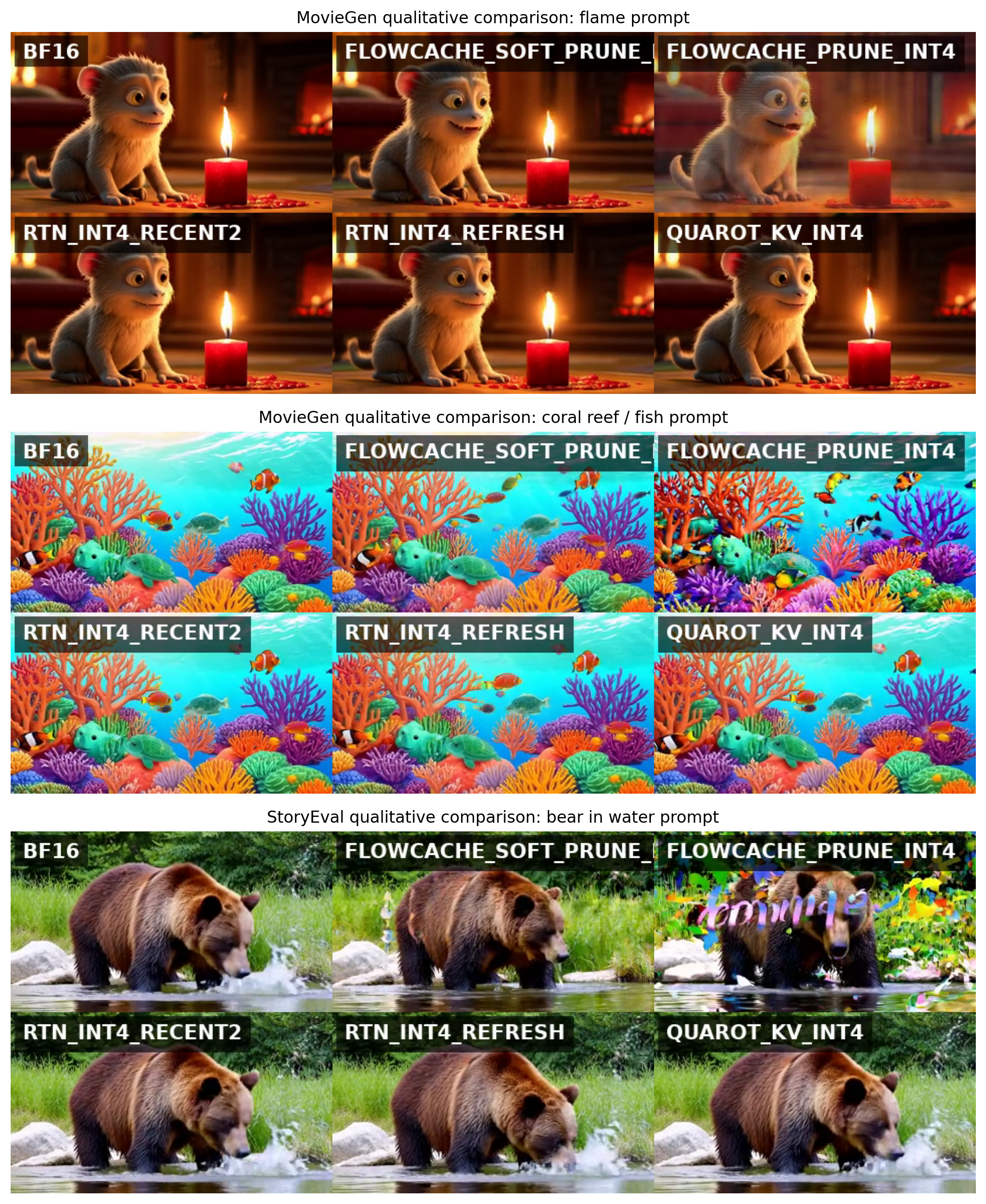}
  \caption{Curated six-method qualitative comparisons for two \benchmark{MovieGen} examples and one \benchmark{StoryEval} example.}
  \label{fig:qualitative}
\end{figure*}

Figure~\ref{fig:qualitative} shows the three prompt families we used most in the live presentation: a candle/flame prompt and a coral reef / fish prompt from \benchmark{MovieGen}, and a bear-in-water prompt from \benchmark{StoryEval}. These panels help explain why realism and fidelity must be separated. The FlowCache-inspired soft-prune INT4 method usually remains visually plausible and often aesthetically strong, but its videos can deviate structurally from BF16. QuaRot and RTN\_RECENT2 often preserve object layout and temporal identity better, albeit without improving realized peak VRAM. The spatially mixed failure case shows the opposite extreme: a method can be explicitly designed to preserve ``important'' regions and still break scene coherence under autoregressive rollout.

\subsection{Why Does Peak VRAM Sometimes Increase Despite Compression?}

One of the most surprising results in the study is that some quantized methods compress the KV cache substantially and still exceed BF16 peak VRAM. The three most important examples are \method{QUAROT\_KV\_INT4}, \method{RTN\_INT4\_RECENT2}, and \method{RTN\_INT4\_REFRESH}. Their benchmark-level summary numbers already show the anomaly. On \benchmark{MovieGen}, BF16 peaks at 19.28\,GB. Yet \method{QUAROT\_KV\_INT4} peaks at 19.98\,GB, \method{RTN\_INT4\_RECENT2} at 21.37\,GB, and \method{RTN\_INT4\_REFRESH} at 22.64\,GB.

Our explanation combines implementation inspection with trace evidence. The current integration stores compressed state, but several methods still reconstruct dense BF16 tensors during attention reads or maintain BF16 regions intentionally. QuaRot adds another layer of overhead because rotated low-bit values must be mapped back through inverse Hadamard-style transforms before attention. The recency-protected RTN variant keeps a recent BF16 tail live while reconstructing the older quantized prefix. The refresh variant is even more expensive because it periodically holds and rewrites large BF16 states during refresh. These are transient systems effects, not proof that compression failed mathematically; Appendix~A summarizes the implementation choices behind these behaviors.

\begin{figure*}[t]
  \centering
  \includegraphics[width=\textwidth]{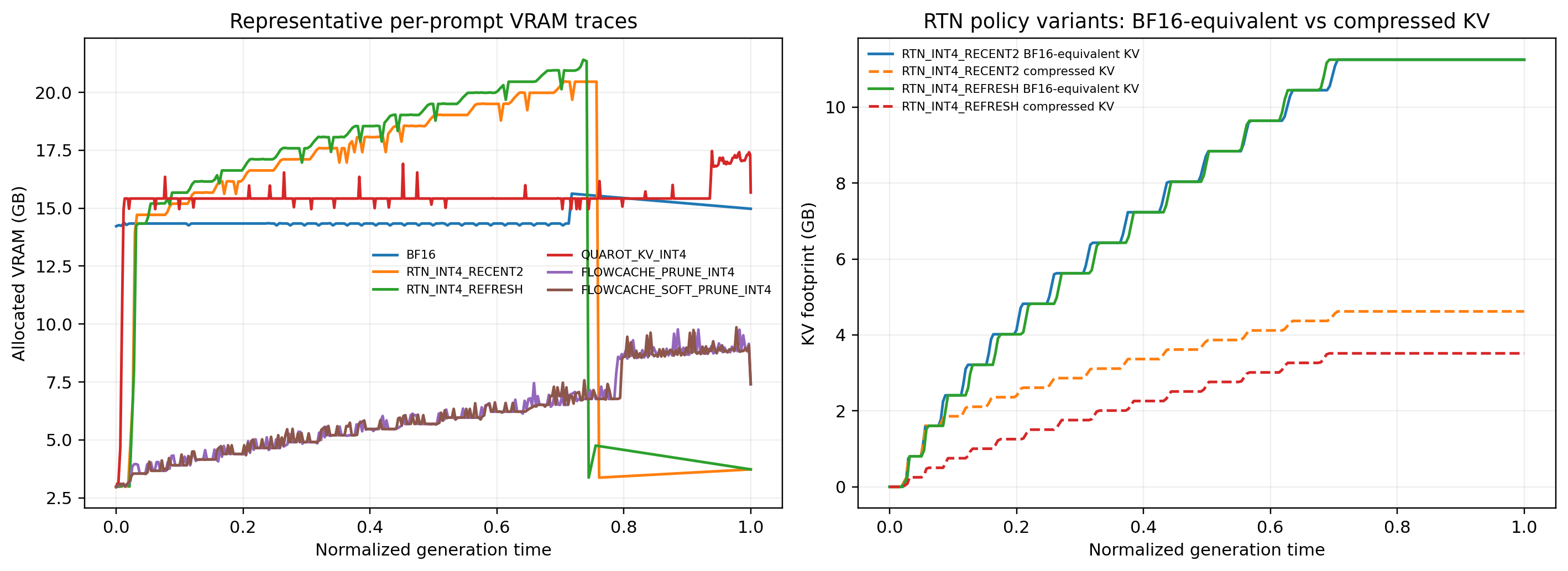}
  \caption{Representative per-prompt traces from available 10-second runs. Left: allocated VRAM over normalized generation time for BF16 and selected quantized methods. Right: BF16-equivalent versus compressed KV footprint for the RTN policy variants. These traces show that some methods genuinely compress the cache but still incur higher measured peaks because the current integration reconstructs or updates large BF16 buffers transiently.}
  \label{fig:traces}
\end{figure*}

The traces in Figure~\ref{fig:traces} make this concrete. In a representative \benchmark{MovieGen} prompt, \method{RTN\_INT4\_REFRESH} reaches an allocated peak of 21.41\,GB even though the active KV cache at that moment corresponds to 11.25\,GB in BF16-equivalent form and only 3.51\,GB in compressed form. \method{RTN\_INT4\_RECENT2} shows the same pattern: 20.47\,GB allocated at peak with 11.25\,GB BF16-equivalent KV and 4.62\,GB compressed KV. The compression is real, but the observed maximum is dominated by reconstruction and update phases rather than by the steady-state compressed footprint. This is precisely why these methods still matter scientifically even when they are not the best deployment choice.

\begin{figure*}[t]
  \centering
  \begin{subfigure}[t]{0.49\textwidth}
    \includegraphics[width=\linewidth]{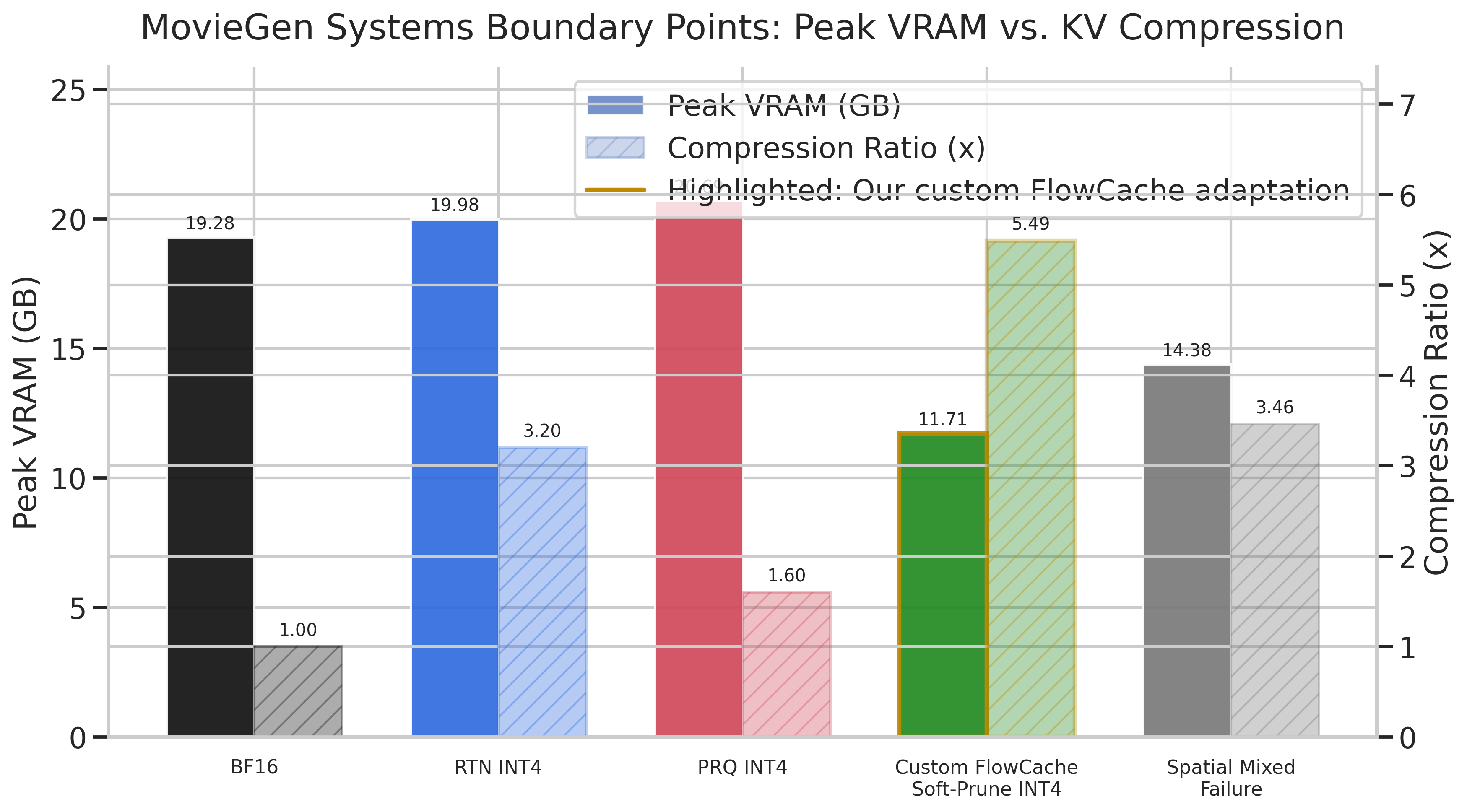}
    \caption{Peak VRAM versus compression.}
  \end{subfigure}
  \hfill
  \begin{subfigure}[t]{0.49\textwidth}
    \includegraphics[width=\linewidth]{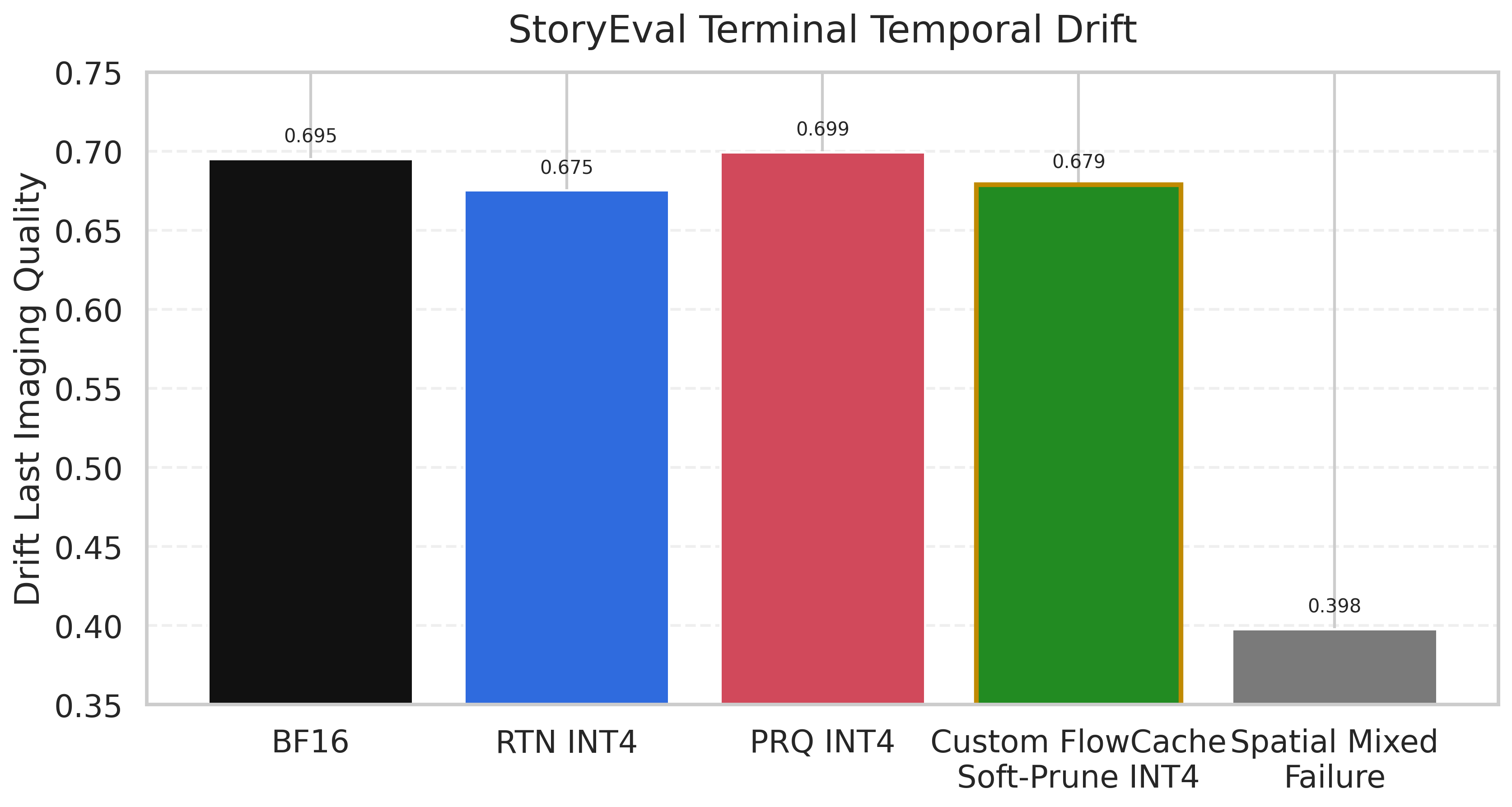}
    \caption{Temporal drift comparison.}
  \end{subfigure}
  \caption{Complementary summary views of systems and long-horizon behavior across methods. Left: peak VRAM versus realized compression highlights which methods convert KV savings into actual memory relief. Right: temporal drift comparison highlights which methods preserve rollout stability as context accumulates.}
  \label{fig:repo-summary-a}
\end{figure*}

Figure~\ref{fig:repo-summary-a} provides a compact cross-method summary of the two most deployment-relevant global views: memory relief as a function of realized compression, and terminal temporal drift across methods.

\clearpage
\section{Discussion}

The study suggests a clear way to interpret the 33-method landscape. There are deployment winners and research winners, and they are not always the same methods. The deployment winner is the FlowCache-inspired soft-prune INT4 region because it is the only region that simultaneously achieves large realized compression, materially lower peak VRAM, and acceptable runtime. The research winners are methods such as \method{PRQ\_INT4}, \method{QUAROT\_KV\_INT4}, and \method{RTN\_INT4\_RECENT2}, which identify algorithmic ideas that preserve quality better but have not yet been turned into the most efficient memory integration.

The negative results are equally important. Spatial mixed precision looked attractive because it encoded a plausible vision prior: preserve foreground objects more carefully than background content. Yet in autoregressive rollout that prior appears misaligned with the actual causal dependencies of the model, and the method collapses badly on both realism and drift. This is a useful warning against over-trusting intuition about what the model should care about.

More broadly, the paper reinforces a methodological point. For long-horizon video generation, reporting only one metric is misleading. Compression ratio without peak VRAM can hide transient-memory failures. VBench realism without BF16-reference fidelity can hide structural hallucination. Fidelity without runtime can recommend impractical methods. A useful empirical study therefore has to treat memory, speed, realism, fidelity, and drift as a joint analysis problem rather than a single-score leaderboard.

\section{Future Work}

The first next step is systems-oriented and follows directly from the peak-memory anomalies in Section~4. The current stack often stores quantized KV state but still reconstructs or preserves dense BF16 buffers during attention, refresh, or recency-protected reads. The cleanest immediate follow-up is therefore to redesign the attention path so compressed KV tensors can be consumed with far less transient BF16 materialization. This is the systems experiment most likely to tell us whether the quality-preserving ideas identified here, especially recency protection and outlier-aware quantization, can become practical deployment winners rather than only research winners.

The second next step is to reproduce newer KV-aware long-video systems on the same benchmark harness. The most relevant current target is Quant VideoGen, which introduces both QVG and the stronger QVG-Pro operating mode for autoregressive video generation \citep{xi2026qvg}. As of March 17, 2026, we did not identify an official public code release linked from the arXiv record, so reproduction on our stack remains an open engineering task rather than a simple re-run. That is precisely why it is valuable: it would let us test whether the reported quality-memory gains survive on the same Self-Forcing \method{Wan2.1} path, hardware profile, and measurement pipeline used throughout this paper.

The third next step is longer-horizon evaluation on stronger compute. Our current clips are already long enough to expose memory pressure, but they are still short-horizon proxies for the more severe consistency failures that appear over 20 seconds or more. Pushing the same study to substantially longer rollouts would let us measure temporal drift, identity breakdown, and scene inconsistency directly rather than infer them from 10-second behavior and prefix-quality curves.

The fourth next step is benchmarking breadth beyond one causal generator. CausVid and HiAR illustrate that long-horizon video generation can be improved by better causal distillation or hierarchical denoising rather than only by cache compression \citep{yin2025causvid,zou2026hiar}. Evaluating the same 33-method design space, or a distilled subset of it, on those alternative stacks would help determine which conclusions are fundamental to KV-cache policy and which ones are specific to Self-Forcing \method{Wan2.1}.

The fifth next step is to move beyond pure text-to-video benchmarks and into first-frame-grounded and embodied settings. These settings put much sharper pressure on identity retention, layout persistence, and action consistency over long horizons. If a method drifts there, the failure is much easier to observe and much more consequential, making them a stronger downstream testbed for the cache-compression ideas studied here.

\section{Conclusion}

We presented a comprehensive empirical study of KV-cache quantization for self-forcing video generation. The study spans 33 methods, two benchmarks, 610 prompt-level observations, and a unified analysis stack that produces both static figures and an interactive dashboard. The headline conclusion is deliberately nuanced. FlowCache-inspired soft pruning is the strongest practical operating point in the current stack because it delivers real peak-VRAM relief and strong perceptual quality. PRQ, QuaRot, and recency-aware RTN variants remain highly valuable because they identify quality-preserving directions that a better systems integration could make practical. And some plausible ideas, especially spatial mixed precision, fail clearly enough to narrow the design space. For long-horizon video generation, that full map is more useful than a single winner.

\clearpage
\bibliographystyle{plainnat}
\bibliography{references}

\clearpage
\appendix
\onecolumn

\section{Method-by-Method Implementation Notes}

Table~\ref{tab:appendix-method-impl} summarizes the code-level implementation choices behind every evaluated method. These notes are derived from the quantizer classes, the generation entrypoint, and the cache-update path in the Self-Forcing integration. They are intended to make the experimental design auditable from the paper itself.

The full codebase, benchmark harness, analysis scripts, and dashboard are available at \url{https://github.com/suraj-ranganath/kv-quant-longhorizon/}.

\tiny
\setlength{\tabcolsep}{2pt}
\begin{longtable}{>{\raggedright\arraybackslash}p{2.6cm} >{\raggedright\arraybackslash}p{1.2cm} >{\raggedright\arraybackslash}p{5.7cm} >{\raggedright\arraybackslash}p{3.8cm}}
\caption{Code-level implementation notes for all 33 evaluated methods. Shared defaults across the harness include blockwise KV quantization at the causal cache boundary with block size 16 when a quantizer is active.\label{tab:appendix-method-impl}}\\
\toprule
Method & Family & Implementation detail & Runtime / code path \\
\midrule
\endfirsthead
\toprule
Method & Family & Implementation detail & Runtime / code path \\
\midrule
\endhead
BF16 & BF16 & No quantizer object; the causal KV cache is stored and read back in native BF16 throughout generation. & No quantized cache-policy hook; acts as the uncompressed reference path. Code: 01\_generate.py; causal\_model.py \\
RTN\_INT2 & RTN & RTNQuantizer with symmetric blockwise quantization of both K and V, block\_size=16, 2-bit packed storage plus per-block scales. & Quantize-on-write every step; dequantize the active cache back to dense BF16 before attention. Code: rtn.py; 01\_generate.py; causal\_model.py \\
RTN\_INT4 & RTN & RTNQuantizer with symmetric blockwise quantization of both K and V, block\_size=16, 4-bit packed storage plus per-block scales. & Quantize-on-write every step; dequantize the active cache back to dense BF16 before attention. Code: rtn.py; 01\_generate.py; causal\_model.py \\
RTN\_K2\_V4 & RTN & RTNQuantizer mixed-precision variant with 2-bit keys and 4-bit values on the same symmetric blockwise path. & Same per-step quantize/dequantize path as standard RTN. Code: rtn.py; 01\_generate.py \\
\shortstack[l]{RTN\_INT4\\RECENT2} & RTN & RTN INT4 quantizer for the old prefix; the newest 2 frame-blocks are carved out and stored separately as BF16 recent\_k/recent\_v state. & Per-step quantize-on-write with recent\_blocks=2; on read, the quantized prefix is dequantized and the BF16 recent tail is copied back in. Code: rtn.py; 01\_generate.py; causal\_model.py \\
\shortstack[l]{RTN\_INT4\\REFRESH} & RTN & RTN INT4 quantizer, but used only during the clean-context refresh pass rather than every denoising step. & quantize\_cadence=refresh\_only; denoising keeps live BF16 cache state, then the timestep-zero context rerun writes quantized cache state. Code: rtn.py; 01\_generate.py; causal\_inference.py \\
KIVI\_INT2 & KIVI & KIVIQuantizer with asymmetric quantization: keys use per-channel scales/zero-points over block tokens, values use per-token scales/zero-points, both at 2 bits. & Per-step quantize-on-write with dense BF16 reconstruction before attention. Code: kivi.py; 01\_generate.py; causal\_model.py \\
KIVI\_INT4 & KIVI & KIVIQuantizer with asymmetric quantization: keys use per-channel scales/zero-points over block tokens, values use per-token scales/zero-points, both at 4 bits. & Per-step quantize-on-write with dense BF16 reconstruction before attention. Code: kivi.py; 01\_generate.py; causal\_model.py \\
KIVI\_K2\_V4 & KIVI & KIVIQuantizer mixed-precision variant with 2-bit key quantization and 4-bit value quantization on the same asymmetric path. & Per-step quantize-on-write with asymmetric dequantization before attention. Code: kivi.py; 01\_generate.py \\
\shortstack[l]{KIVI\_INT4\\REFRESH} & KIVI & KIVI INT4 quantizer invoked only during refresh, keeping the asymmetric K/V quantizer but changing the cache cadence. & refresh\_only cadence during denoising, followed by clean-context re-quantization. Code: kivi.py; 01\_generate.py; causal\_inference.py \\
QUAROT\_KV\_INT2 & QUAROT & QuaRotKVQuantizer applies a Hadamard transform on the channel axis, uses RTN-style symmetric block quantization in rotated space, then inverse-rotates on read at 2 bits. & Per-step quantize-on-write with dense BF16 reconstruction plus inverse Hadamard rotation before attention. Code: quarot\_kv.py; 01\_generate.py; causal\_model.py \\
QUAROT\_KV\_INT4 & QUAROT & QuaRotKVQuantizer applies a Hadamard transform on the channel axis, uses RTN-style symmetric block quantization in rotated space, then inverse-rotates on read at 4 bits. & Per-step quantize-on-write with dense BF16 reconstruction plus inverse Hadamard rotation before attention. Code: quarot\_kv.py; 01\_generate.py; causal\_model.py \\
\shortstack[l]{QUAROT\_KV\_INT4\\RECENT2} & QUAROT & QuaRot INT4 on the old prefix plus a BF16 recent tail that skips rotation/quantization for the newest 2 frame-blocks. & Per-step quantize-on-write with recent\_blocks=2; rotated old context is dequantized, inverse-rotated, and then concatenated with the BF16 tail. Code: quarot\_kv.py; 01\_generate.py; causal\_model.py \\
\shortstack[l]{QUAROT\_KV\_INT4\\REFRESH} & QUAROT & QuaRot INT4 quantizer combined with refresh-only cache cadence rather than per-step writes. & BF16 cache persists during denoising; a clean-context refresh pass performs rotated quantization and cache write-back. Code: quarot\_kv.py; 01\_generate.py; causal\_inference.py \\
PRQ\_INT2 & PRQ & PRQQuantizer uses two-stage symmetric residual coding: stage 1 quantizes the original tensor, stage 2 quantizes the reconstruction residual in 64-block chunks, with base storage at 2 bits. & Per-step quantize-on-write; read path reconstructs both stages and sums them before attention. Code: prq.py; 01\_generate.py \\
PRQ\_INT4 & PRQ & PRQQuantizer uses two-stage symmetric residual coding: stage 1 quantizes the original tensor, stage 2 quantizes the reconstruction residual in 64-block chunks, with base storage at 4 bits. & Per-step quantize-on-write; read path reconstructs both stages and sums them before attention. Code: prq.py; 01\_generate.py \\
QAQ\_INT2 & QAQ & QAQQuantizer clips each block to an outlier threshold, quantizes the clipped bulk asymmetrically at 2 bits, and stores explicit indices/values for outliers in higher precision. & Per-step quantize-on-write; dequantization restores the bulk and then writes preserved outliers back into the tensor. Code: qaq.py; 01\_generate.py \\
QAQ\_INT4 & QAQ & QAQQuantizer clips each block to an outlier threshold, quantizes the clipped bulk asymmetrically at 4 bits, and stores explicit indices/values for outliers in higher precision. & Per-step quantize-on-write; dequantization restores the bulk and then writes preserved outliers back into the tensor. Code: qaq.py; 01\_generate.py \\
AGE\_TIER\_INT2 & AGE\_TIER & AgeTierQuantizer splits the sequence by a trailing recent\_ratio mask, applies a higher-bit recent quantizer to the recent slice, and applies an 2-bit old quantizer to the older slice. & Per-step quantize-on-write; both slices are dequantized separately and stitched back together before attention. Code: age\_tier.py; 01\_generate.py \\
AGE\_TIER\_INT4 & AGE\_TIER & AgeTierQuantizer splits the sequence by a trailing recent\_ratio mask, applies a higher-bit recent quantizer to the recent slice, and applies an 4-bit old quantizer to the older slice. & Per-step quantize-on-write; both slices are dequantized separately and stitched back together before attention. Code: age\_tier.py; 01\_generate.py \\
TPTQ\_INT2 & TPTQ & TPTQQuantizer keeps a recent slice with a higher-precision recent quantizer, compresses older tokens with PRQQuantizer, and separately preserves the strongest old-key outliers up to outlier\_max\_ratio. & Per-step quantize-on-write; the old PRQ reconstruction and stored outliers are merged on read. Code: tptq.py; prq.py; 01\_generate.py \\
\shortstack[l]{FLOWCACHE\_HYBRID\\INT2} & FLOWCACHE & FlowCacheHybridQuantizer partitions the active cache into frame-aligned chunks, keeps recent chunks at higher precision, compresses older chunks at low precision, and modulates the recent budget by layer role. & Per-step quantize-on-write using chunk boundaries from frame\_seq\_length and layer budget scaling. Code: flowcache\_hybrid.py; 01\_generate.py \\
\shortstack[l]{FLOWCACHE\_ADAPTIVE\\INT2} & FLOWCACHE & FlowCacheAdaptiveQuantizer extends Hybrid by computing chunk summaries, scoring old chunks by relative-L1 delta and recency, and upgrading a small important-old subset to higher precision. & Per-step quantize-on-write with adaptive importance scoring at each layer. Code: flowcache\_adaptive.py; flowcache\_hybrid.py; 01\_generate.py \\
\shortstack[l]{FLOWCACHE\_PRUNE\\INT2} & FLOWCACHE & FlowCachePruneQuantizer keeps recent chunks and important old chunks, retains only a limited low-bit old subset at 2 bits, and treats the rest as pruned chunks reconstructed as zeros. & Chunk plans are cached and refreshed only when frame-aligned chunk count changes; dequantization leaves pruned spans zero-filled. Code: flowcache\_prune.py; flowcache\_adaptive.py; 01\_generate.py \\
\shortstack[l]{FLOWCACHE\_PRUNE\\INT4} & FLOWCACHE & FlowCachePruneQuantizer keeps recent chunks and important old chunks, retains only a limited low-bit old subset at 4 bits, and treats the rest as pruned chunks reconstructed as zeros. & Chunk plans are cached and refreshed only when frame-aligned chunk count changes; dequantization leaves pruned spans zero-filled. Code: flowcache\_prune.py; flowcache\_adaptive.py; 01\_generate.py \\
\shortstack[l]{FLOWCACHE\_SOFT\\PRUNE\_INT2} & FLOWCACHE & FlowCacheSoftPruneQuantizer follows the prune plan but stores one pooled BF16 summary per pruned chunk and repeats that summary token across the evicted span during reconstruction, with old retained chunks at 2 bits. & Same cached chunk-plan logic as hard prune; pruned spans are reconstructed from summary tokens instead of zeros. Code: flowcache\_soft\_prune.py; flowcache\_prune.py; 01\_generate.py \\
\shortstack[l]{FLOWCACHE\_SOFT\\PRUNE\_INT4} & FLOWCACHE & FlowCacheSoftPruneQuantizer follows the prune plan but stores one pooled BF16 summary per pruned chunk and repeats that summary token across the evicted span during reconstruction, with old retained chunks at 4 bits. & Same cached chunk-plan logic as hard prune; pruned spans are reconstructed from summary tokens instead of zeros. Code: flowcache\_soft\_prune.py; flowcache\_prune.py; 01\_generate.py \\
FLOWCACHE\_NATIVE & FLOWCACHE & No KV quantizer; instead, a FlowCacheReuseManager is attached to the pipeline and decides when internal features can be reused based on a relative-L1 drift threshold and optional warmup. & Latency-oriented runtime reuse path rather than memory compression; representative runs use flowcache\_native\_rel\_l1\_thresh metadata in the registry. Code: 01\_generate.py::attach\_flowcache\_native \\
\shortstack[l]{FLOWCACHE\_NATIVE\\SOFT\_PRUNE\_INT4} & FLOWCACHE & Combines FlowCacheReuseManager with a FlowCacheSoftPrune INT4 quantizer so generation can skip recomputation on low-drift steps while still compressing retained cache chunks. & Reuse-manager gating plus the soft-prune chunk planner; this is the only method that composes latency reuse with cache compression in one path. Code: 01\_generate.py::attach\_flowcache\_native; flowcache\_soft\_prune.py \\
\shortstack[l]{SPATIAL\_MIXED\\FG\_RTN\_INT4\\BG\_RTN\_INT2} & SPATIAL\_MIXED & SpatialMixedQuantizer builds a foreground/background token mask from temporal variance and routes foreground tokens to RTN\_INT4 while routing background tokens to RTN\_INT2. & Per-step quantize-on-write; mask\_policy and foreground-ratio bounds decide whether the mask is threshold-based, top-k, or hybrid. Code: spatial\_mixed.py; 01\_generate.py \\
\shortstack[l]{SPATIAL\_MIXED\\FG\_RTN\_INT4\\BG\_RTN\_INT4} & SPATIAL\_MIXED & SpatialMixedQuantizer builds a foreground/background token mask from temporal variance and routes foreground tokens to RTN\_INT4 while routing background tokens to RTN\_INT4. & Per-step quantize-on-write; mask\_policy and foreground-ratio bounds decide whether the mask is threshold-based, top-k, or hybrid. Code: spatial\_mixed.py; 01\_generate.py \\
\shortstack[l]{SPATIAL\_MIXED\\FG\_KIVI\_INT4\\BG\_KIVI\_INT2} & SPATIAL\_MIXED & SpatialMixedQuantizer builds a foreground/background token mask from temporal variance and routes foreground tokens to KIVI\_INT4 while routing background tokens to KIVI\_INT2. & Per-step quantize-on-write; mask\_policy and foreground-ratio bounds decide whether the mask is threshold-based, top-k, or hybrid. Code: spatial\_mixed.py; 01\_generate.py \\
\shortstack[l]{SPATIAL\_MIXED\\FG\_QUAROT\_KV\_INT4\\BG\_RTN\_INT2} & SPATIAL\_MIXED & SpatialMixedQuantizer builds a foreground/background token mask from temporal variance and routes foreground tokens to QUAROT\_KV\_INT4 while routing background tokens to RTN\_INT2. & Per-step quantize-on-write; mask\_policy and foreground-ratio bounds decide whether the mask is threshold-based, top-k, or hybrid. Code: spatial\_mixed.py; 01\_generate.py \\
\bottomrule
\end{longtable}

\clearpage
\section{Additional Repository Figures}

\begin{figure}[H]
  \centering
  \includegraphics[width=0.86\textwidth]{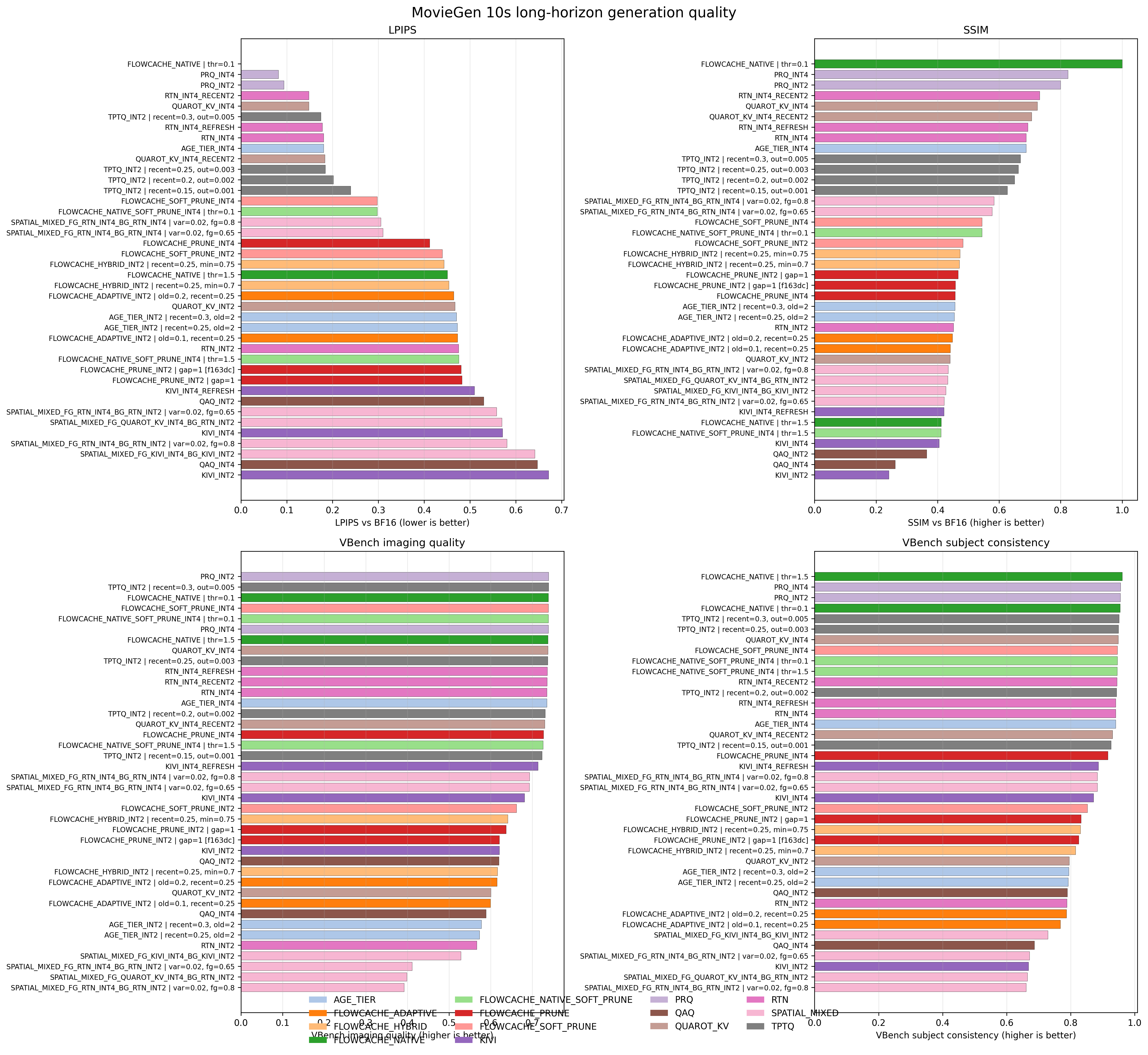}
  \caption{MovieGen generation-quality summary from the repository assets. This plot complements the main-paper realism/fidelity discussion by showing the same benchmark slice in a compact static view.}
  \label{fig:repo-moviegen-generation-quality}
\end{figure}

\begin{figure}[H]
  \centering
  \includegraphics[width=0.86\textwidth]{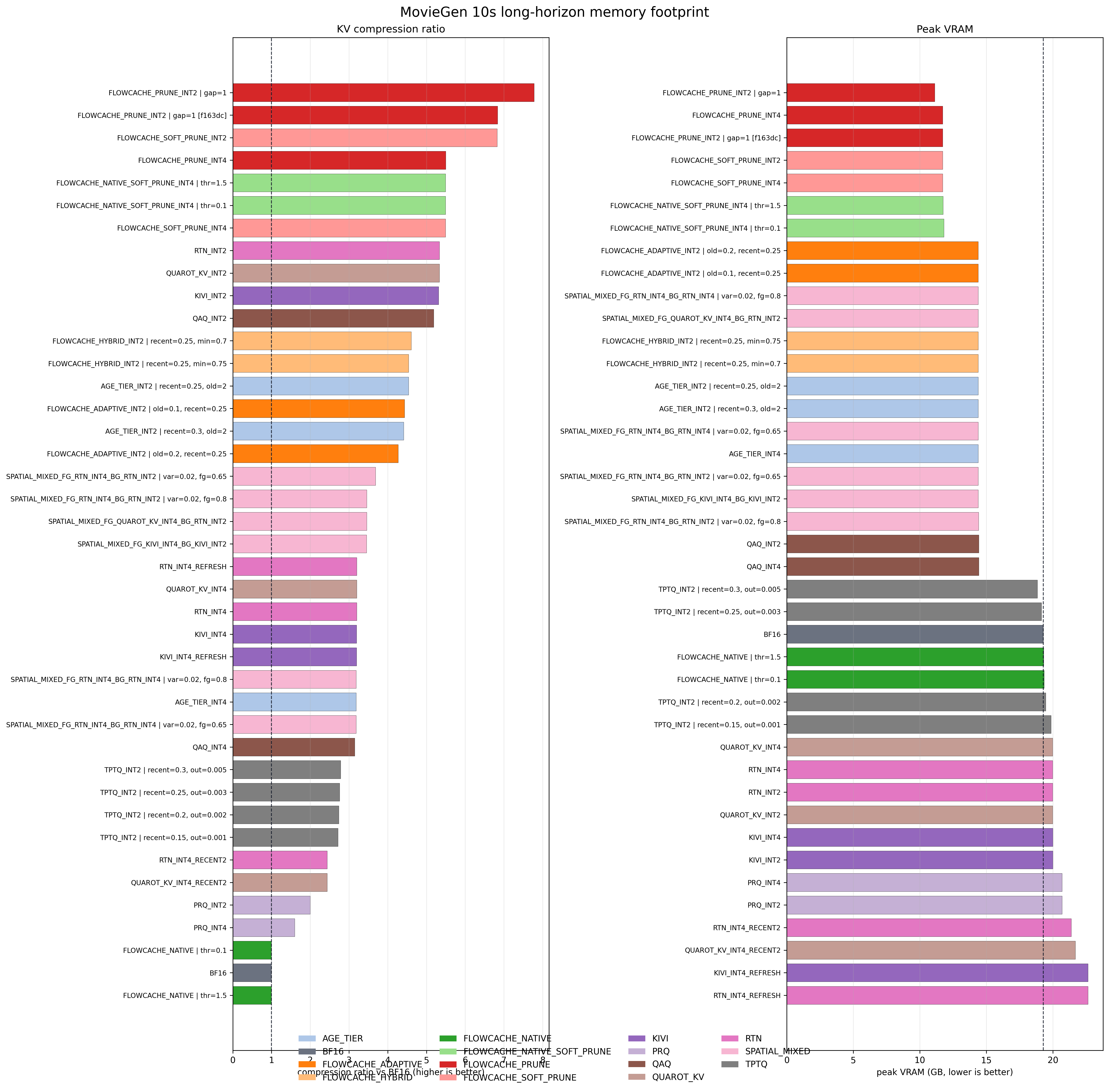}
  \caption{MovieGen memory-footprint summary from the repository assets. This view complements the main-paper peak-VRAM and compression analysis with the dashboard-style static summary used in presentation material.}
  \label{fig:repo-moviegen-memory}
\end{figure}

\begin{figure}[H]
  \centering
  \includegraphics[width=0.86\textwidth]{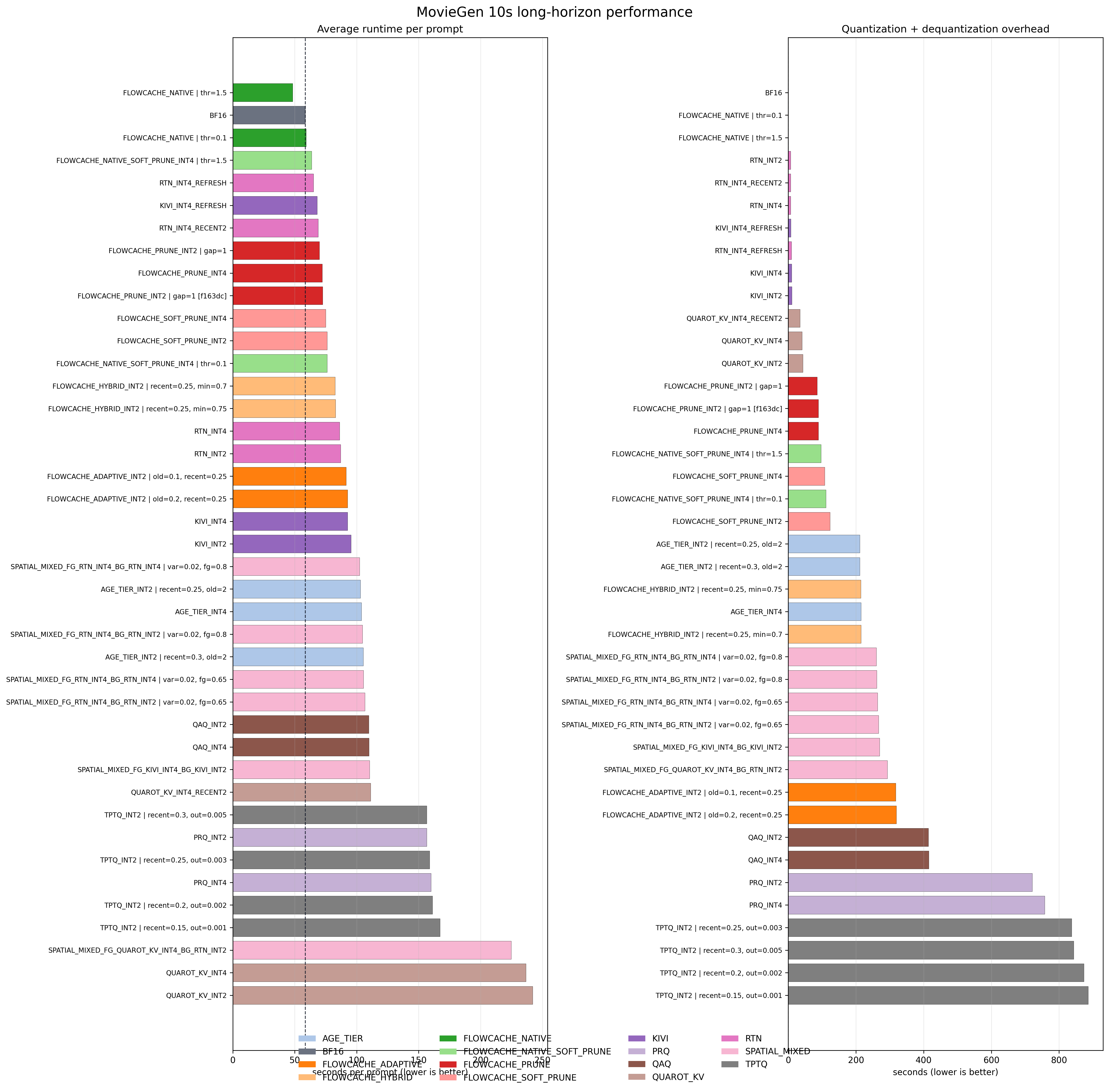}
  \caption{MovieGen performance summary from the repository assets. We include it separately so runtime and deployment-facing behavior remain legible in print.}
  \label{fig:repo-moviegen-performance}
\end{figure}

\clearpage
\section{Full Benchmark Tables}

\noindent
Appendix tables use compact human-readable method labels for readability. The exact raw method identifiers are preserved in the repository datasets and dashboard exports.

\scriptsize
\setlength{\tabcolsep}{3pt}
\begin{longtable}{>{\raggedright\arraybackslash}p{5.0cm}rrrrrrrr}
\caption{Moviegen benchmark summary for all methods. BF16 is the reference baseline row. Img. denotes imaging quality, and Drift denotes drift-last imaging quality.}\\
\toprule
Method & Comp. & Peak VRAM & Runtime & Img. & SSIM & LPIPS & PSNR & Drift \\
\midrule
\endfirsthead
\toprule
Method & Comp. & Peak VRAM & Runtime & Img. & SSIM & LPIPS & PSNR & Drift \\
\midrule
\endhead
FlowCache Prune INT2 & 7.78 & 11.11 & 69.9 & 0.637 & 0.467 & 0.483 & 15.26 & 0.633 \\
FlowCache Soft-Prune INT2 & 6.82 & 11.71 & 76.1 & 0.662 & 0.482 & 0.440 & 15.84 & 0.658 \\
FlowCache Prune INT4 & 5.50 & 11.71 & 72.2 & 0.727 & 0.457 & 0.412 & 15.30 & 0.726 \\
FlowCache Soft-Prune INT4 & 5.49 & 11.71 & 75.0 & 0.739 & 0.544 & 0.297 & 17.67 & 0.738 \\
FlowCache Native Soft-Prune INT4 & 5.49 & 11.74 & 63.6 & 0.726 & 0.411 & 0.475 & 13.26 & 0.724 \\
RTN INT2 & 5.33 & 19.98 & 87.1 & 0.567 & 0.451 & 0.475 & 15.04 & 0.562 \\
QuaRot KV INT2 & 5.33 & 19.98 & 242.0 & 0.601 & 0.440 & 0.467 & 14.73 & 0.597 \\
KIVI INT2 & 5.31 & 19.99 & 95.5 & 0.621 & 0.241 & 0.671 & 11.42 & 0.618 \\
QAQ INT2 & 5.18 & 14.42 & 109.8 & 0.620 & 0.365 & 0.530 & 13.34 & 0.618 \\
FlowCache Hybrid INT2 & 4.61 & 14.38 & 82.6 & 0.616 & 0.471 & 0.454 & 15.62 & 0.612 \\
Age-Tier INT2 & 4.41 & 14.38 & 105.3 & 0.578 & 0.457 & 0.470 & 15.18 & 0.573 \\
FlowCache Adaptive INT2 & 4.27 & 14.38 & 92.6 & 0.616 & 0.448 & 0.464 & 15.19 & 0.611 \\
SpatialMixed fg RTN INT 4 / bg RTN INT 2 & 3.68 & 14.38 & 106.6 & 0.411 & 0.421 & 0.558 & 13.93 & 0.407 \\
Spatial Mixed (QuaRot fg, RTN bg) & 3.46 & 14.38 & 224.8 & 0.399 & 0.433 & 0.570 & 14.06 & 0.394 \\
SpatialMixed fg KIVI INT 4 / bg KIVI INT 2 & 3.45 & 14.38 & 110.4 & 0.529 & 0.427 & 0.642 & 13.72 & 0.521 \\
QuaRot KV INT4 & 3.20 & 19.98 & 236.6 & 0.738 & 0.724 & 0.148 & 22.64 & 0.738 \\
RTN INT4 Refresh & 3.20 & 22.64 & 65.0 & 0.736 & 0.693 & 0.178 & 21.45 & 0.735 \\
RTN INT4 & 3.20 & 19.98 & 86.3 & 0.735 & 0.688 & 0.180 & 21.32 & 0.734 \\
KIVI INT4 Refresh & 3.19 & 22.63 & 68.1 & 0.714 & 0.420 & 0.509 & 13.73 & 0.712 \\
KIVI INT4 & 3.19 & 19.99 & 92.7 & 0.681 & 0.405 & 0.571 & 13.07 & 0.678 \\
Age-Tier INT4 & 3.18 & 14.38 & 103.9 & 0.735 & 0.688 & 0.180 & 21.32 & 0.734 \\
SpatialMixed fg RTN INT 4 / bg RTN INT 4 & 3.18 & 14.38 & 105.5 & 0.693 & 0.577 & 0.310 & 18.89 & 0.690 \\
QAQ INT4 & 3.14 & 14.42 & 110.0 & 0.589 & 0.262 & 0.647 & 11.97 & 0.586 \\
TPTQ INT2 & 2.72 & 19.85 & 167.2 & 0.724 & 0.627 & 0.240 & 19.91 & 0.722 \\
RTN INT4 Recent2 & 2.43 & 21.37 & 68.9 & 0.736 & 0.732 & 0.148 & 23.69 & 0.735 \\
QuaRot KV INT4 Recent2 & 2.43 & 21.69 & 111.3 & 0.730 & 0.706 & 0.183 & inf & 0.729 \\
PRQ INT2 & 2.00 & 20.69 & 156.6 & 0.739 & 0.800 & 0.094 & 25.13 & 0.740 \\
PRQ INT4 & 1.60 & 20.69 & 160.0 & 0.739 & 0.824 & 0.082 & 26.54 & 0.739 \\
BF16 & 1.00 & 19.28 & 58.6 & 0.739 & 1.000 & 0.000 & inf & 0.739 \\
FlowCache Native & 1.00 & 19.31 & 48.3 & 0.738 & 0.412 & 0.451 & 13.25 & 0.737 \\
QuaRot KV INT4 Refresh & -- & 22.82 & 97.5 & 0.722 & 0.613 & 0.214 & 19.64 & 0.719 \\
RTN K2/V4 & -- & 22.68 & 75.3 & 0.531 & 0.434 & 0.495 & 14.74 & 0.524 \\
KIVI K2/V4 & -- & 22.67 & 76.3 & 0.623 & 0.374 & 0.578 & 13.03 & 0.619 \\
\bottomrule
\end{longtable}

\begin{longtable}{>{\raggedright\arraybackslash}p{5.0cm}rrrrrrrr}
\caption{Storyeval benchmark summary for all methods. BF16 is the reference baseline row. Img. denotes imaging quality, and Drift denotes drift-last imaging quality.}\\
\toprule
Method & Comp. & Peak VRAM & Runtime & Img. & SSIM & LPIPS & PSNR & Drift \\
\midrule
\endfirsthead
\toprule
Method & Comp. & Peak VRAM & Runtime & Img. & SSIM & LPIPS & PSNR & Drift \\
\midrule
\endhead
FlowCache Prune INT2 & 7.68 & 11.14 & 70.2 & 0.516 & 0.492 & 0.506 & 14.26 & 0.516 \\
FlowCache Soft-Prune INT2 & 6.72 & 11.76 & 74.4 & 0.532 & 0.497 & 0.495 & 14.18 & 0.536 \\
FlowCache Prune INT4 & 5.43 & 11.75 & 72.4 & 0.682 & 0.465 & 0.490 & inf & 0.680 \\
FlowCache Soft-Prune INT4 & 5.42 & 11.76 & 75.2 & 0.680 & 0.518 & 0.416 & inf & 0.679 \\
FlowCache Native Soft-Prune INT4 & 5.42 & 11.78 & 64.2 & 0.657 & 0.436 & 0.523 & 11.94 & 0.657 \\
RTN INT2 & 5.33 & 19.98 & 86.1 & 0.464 & 0.459 & 0.526 & 13.57 & 0.471 \\
QuaRot KV INT2 & 5.33 & 19.98 & 239.0 & 0.477 & 0.443 & 0.532 & 13.01 & 0.480 \\
KIVI INT2 & 5.31 & 19.99 & 94.7 & 0.531 & 0.243 & 0.735 & inf & 0.527 \\
QAQ INT2 & 5.19 & 14.42 & 109.9 & 0.579 & 0.324 & 0.635 & 11.79 & 0.585 \\
FlowCache Hybrid INT2 & 4.59 & 14.38 & 82.2 & 0.492 & 0.479 & 0.512 & 13.95 & 0.494 \\
Age-Tier INT2 & 4.41 & 14.38 & 101.9 & 0.469 & 0.463 & 0.523 & 13.67 & 0.473 \\
FlowCache Adaptive INT2 & 4.26 & 14.38 & 91.3 & 0.498 & 0.451 & 0.528 & inf & 0.496 \\
SpatialMixed fg RTN INT 4 / bg RTN INT 2 & 3.69 & 14.38 & 106.6 & 0.421 & 0.430 & 0.587 & 12.64 & 0.419 \\
Spatial Mixed (QuaRot fg, RTN bg) & 3.46 & 14.38 & 224.1 & 0.400 & 0.444 & 0.599 & 12.97 & 0.398 \\
SpatialMixed fg KIVI INT 4 / bg KIVI INT 2 & 3.45 & 14.38 & 110.2 & 0.454 & 0.453 & 0.654 & 13.20 & 0.451 \\
QuaRot KV INT4 & 3.20 & 19.98 & 239.6 & 0.687 & 0.685 & 0.217 & 19.25 & 0.689 \\
RTN INT4 & 3.20 & 19.98 & 88.8 & 0.674 & 0.661 & 0.245 & 18.66 & 0.675 \\
RTN INT4 Refresh & 3.20 & 22.64 & 64.6 & 0.678 & 0.654 & 0.252 & 18.55 & 0.679 \\
KIVI INT4 & 3.19 & 19.99 & 93.0 & 0.635 & 0.424 & 0.575 & 12.41 & 0.635 \\
KIVI INT4 Refresh & 3.19 & 22.63 & 66.7 & 0.645 & 0.414 & 0.569 & 12.49 & 0.641 \\
Age-Tier INT4 & 3.18 & 14.38 & 102.4 & 0.674 & 0.661 & 0.245 & 18.66 & 0.676 \\
SpatialMixed fg RTN INT 4 / bg RTN INT 4 & 3.18 & 14.38 & 106.6 & 0.606 & 0.587 & 0.349 & inf & 0.607 \\
QAQ INT4 & 3.15 & 14.42 & 109.9 & 0.567 & 0.238 & 0.719 & 11.03 & 0.571 \\
TPTQ INT2 & 2.72 & 19.77 & 166.6 & 0.654 & 0.615 & 0.301 & 17.03 & 0.658 \\
RTN INT4 Recent2 & 2.43 & 21.37 & 68.6 & 0.680 & 0.721 & 0.187 & inf & 0.684 \\
QuaRot KV INT4 Recent2 & 2.43 & 21.69 & 112.9 & 0.666 & 0.707 & 0.222 & inf & 0.670 \\
PRQ INT2 & 2.00 & 20.69 & 155.6 & 0.698 & 0.733 & 0.179 & 20.66 & 0.698 \\
PRQ INT4 & 1.60 & 20.69 & 158.0 & 0.699 & 0.724 & 0.188 & inf & 0.699 \\
BF16 & 1.00 & 19.28 & 56.8 & 0.693 & 1.000 & 0.000 & inf & 0.695 \\
FlowCache Native & 1.00 & 19.31 & 49.0 & 0.681 & 0.451 & 0.508 & 11.95 & 0.682 \\
\bottomrule
\end{longtable}

\end{document}